\documentclass[letterpaper, 10 pt, conference]{ieeeconf}  

\IEEEoverridecommandlockouts                              %
\usepackage[utf8]{inputenc}
\usepackage[T1]{fontenc}
\usepackage{float}
\usepackage{graphicx}
\usepackage{booktabs}

\usepackage{adjustbox}
\usepackage{textcomp}
\usepackage{array,multirow,graphicx}

\usepackage{etoolbox}
\makeatletter
\patchcmd{\@makecaption}
  {\scshape}
  {}
  {}
  {}
\makeatletter
\patchcmd{\@makecaption}
  {\\}
  {.\ }
  {}
  {}
\makeatother

\usepackage{stfloats}

\title{\LARGE \bf
DensePASS: Dense Panoramic Semantic Segmentation via Unsupervised Domain Adaptation with Attention-Augmented Context Exchange
}
\author{Chaoxiang Ma$^{1}$, Jiaming Zhang$^{1}$, Kailun Yang$^{1}$, Alina Roitberg$^{1}$ and Rainer Stiefelhagen$^{1}$
\thanks{This work was supported in part through the AccessibleMaps project by the Federal Ministry of Labor and Social Affairs (BMAS) under the Grant No. 01KM151112, in part by the University of Excellence through the ``KIT Future Fields'' project, and in part by Hangzhou SurImage Company Ltd.
\textit{(Corresponding author: Kailun Yang.)}}
\thanks{$^{1}$Authors are with Institute for Anthropomatics and Robotics, Karlsruhe Institute of Technology, Germany (e-mail: chaoxiang.ma.1024@gmail.com, \{jiaming.zhang, kailun.yang, alina.roitberg, rainer.stiefelhagen\}@kit.edu).}
\thanks{Code and dataset will be made publicly available at: https://github.com/chma1024/DensePASS}
}
\begin{document}

\maketitle
\thispagestyle{empty}
\pagestyle{empty}

\begin{abstract}

Intelligent vehicles clearly benefit from the expanded Field of View (FoV) of the $360^\circ$ sensors, but the vast majority of available semantic segmentation training images are captured with pinhole cameras. In this work, we look at this problem through the lens of domain adaptation and bring \emph{panoramic} semantic segmentation to a setting, where labelled training data originates from a different distribution of conventional \emph{pinhole} camera images. First, we formalize the task of unsupervised domain adaptation for panoramic semantic segmentation, where a network trained on labelled examples from the \emph{source domain} of pinhole camera data is deployed in a different \emph{target domain} of panoramic images, for which no labels are available. To validate this idea, we collect and publicly release \textsc{DensePASS} - a novel densely annotated dataset for panoramic segmentation under cross-domain conditions, specifically built to study the \textsc{Pinhole$\rightarrow$Panoramic} transfer and accompanied with pinhole camera training examples obtained from Cityscapes. \textsc{DensePASS} covers both, labelled- and unlabelled $360^\circ$ images, with the labelled data comprising $19$ classes which explicitly fit the categories available in the source domain (\textit{i.e.} pinhole) data. To meet the challenge of domain shift, we leverage the current progress of attention-based mechanisms and build a generic framework for cross-domain panoramic semantic segmentation based on different variants of attention-augmented domain adaptation modules. Our framework facilitates information exchange at local- and global levels when learning the domain correspondences and improves the domain adaptation performance of two standard segmentation networks by $6.05\%$ and $11.26\%$ in Mean IoU.

\end{abstract}

\section{INTRODUCTION}

Semantic segmentation is essential for perception of intelligent vehicles as it enables to locate key entities of a driving scene, such as \emph{road}, \emph{sidewalk} or \emph{person}, by assigning a category label to every image pixel. 
While semantic segmentation results have increased at a rapid pace since the emergence of fully convolutional networks~\cite{fcn}, most of the previous frameworks are developed under the assumption that the images are captured with a \emph{pinhole} camera.
However, the comparably narrow Field of View~(FoV)  largely limits the perception capacity and can be addressed by mounting multiple sensors, which, in return, requires additional mechanisms for data fusion~\cite{restricted}.
Recently, leveraging a single \emph{panoramic} camera, which offers a unified $360^\circ$ perception of the driving environment,  started to gain attention as a novel alternative for expanding the FoV~\cite{pass}.

\begin{figure}[!t]
  \centering
  \includegraphics[width=0.485\textwidth]{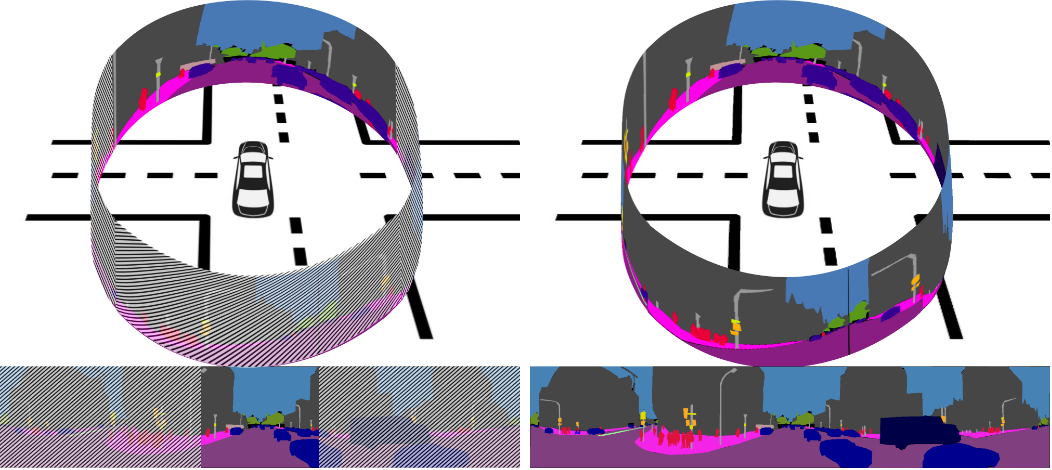}
  \vskip-2ex
  \caption{FoV comparison between pinhole forward-view and $360^\circ$ panoramic surround-view imaging of self-driving road scenes.}
  \label{fig.3d}
  \vskip-2ex
\end{figure}

\begin{figure}[!t]
  \centering
  \includegraphics[width=0.485\textwidth]{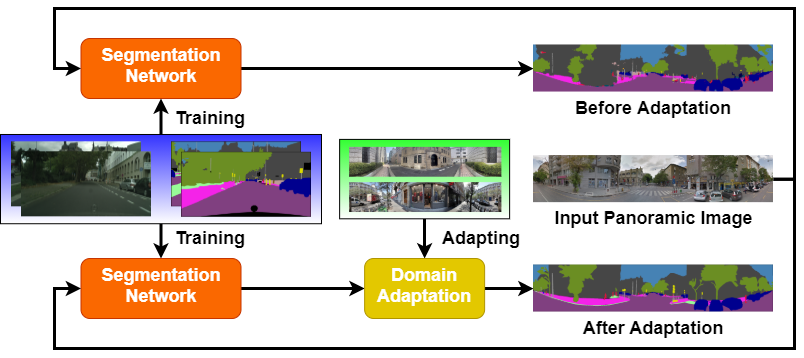}
  \vskip-2ex
  \caption{An overview of the formalized task of domain adaptation for panoramic semantic segmentation. The \emph{source} domain dataset (blue background) contains \emph{pinhole} images with semantic annotations, while the \emph{target} dataset (green background) contains \emph{panoramic} images without annotations.}
  \label{fig.introduce}
  \vskip-3ex
\end{figure}

Unfortunately, the scarcity of pixel-wise annotation of panoramic images still hinders the progress in the semantic segmentation community.
At the same time, \emph{domain adaptation} becomes an increasingly popular research topic, giving a new perspective to complement the insufficient coverage of training data in driving scenarios, \textit{e.g.}, at the nighttime~\cite{bridging} or in accident scenes~\cite{issafe}. 
In this paper, we argue that panoramic segmentation might  strongly benefit from the significantly larger datasets available in the domain of standard image segmentation and explore domain adaptation techniques for such knowledge transfer. 
Thereby, we formalize the task of unsupervised domain adaptation for panoramic segmentation in a novel Dense PAnoramic Semantic Segmentation (DensePASS) benchmark, where images in the label-scarce panoramic \emph{target} domain are handled by adapting from data of the label-rich pinhole \emph{source} domain. 

We aim to promote  research  of \emph{panoramic  semantic segmentation under cross-domain conditions} and introduce DensePASS -- a new dataset  with panoramic images collected  all around the globe to encourage  diversity.
To enable credible quantitative evaluation, our benchmark comprises (1) an unlabelled panoramic training set used for optimization of the domain adaptation model and (2) a panoramic test set manually labelled with $19$ classes defined in accordance to Cityscapes~\cite{cityscapes}, a dataset with pinhole images which we use as training data in the label-rich source domain. 
Yet, a straightforward transfer of models trained on pinhole images to panoramic data often results in a significant performance drop, since the layout of the panoramic camera images  passed through the equirectangular projection strongly differs from the standard pinhole camera data.
For example, as shown in Fig.~\ref{fig.3d}, panoramic images have longer horizontal distribution or geometric distortion on both sides of the viewing direction, resulting in a considerable domain shift. 

To effectively utilize label-rich pinhole image datasets~\cite{cityscapes,wilddash} for label-scarce panoramic segmentation, we systematically examine different Domain Adaptation~(DA) strategies for the \emph{Pinhole to Panoramic Domain Adaptation~(P2PDA)} (the formalized P2PDA task illustrated in Fig.~\ref{fig.introduce}).
Furthermore, we implement a  P2PDA framework with different DA mechanisms: (i) \emph{Segmentation Domain Adaptation Module~(SDAM)}, (ii) \emph{Attentional Domain Adaptation Module~(ADAM)}, and (iii) \emph{Regional Context Domain Adaptation Module~(RCDAM)}.
The proposed \emph{SDAM} module allows greater flexibility than the previous DA methods~\cite{adaptsegnet,all_about_structure} used in the output space as it can be plugged in at different feature levels. 
Another challenge is learning good feature representations, which is not only distinguishing various categories with similar appearances, but also linking the same category at diverse locations across the $360^\circ$.
To address this, we leverage the progress of attention-based models~\cite{nonlocal,danet,fanet} and propose the \emph{ADAM} module for capturing long-range dependencies and positional relations.
Lastly, the \emph{RCDAM} module addresses the horizontal distribution of panoramic images and obtains region-level context in order to effectively resist the geometric distortion caused by the equirectangular projection. 
Extensive experiments demonstrate the effectiveness of our framework, exceeding more than 15 state-of-the-art semantic segmentation methods.

In summary, our main contributions are as following:
\begin{itemize}
\item We create and publicly release DensePASS -- a new benchmark for panoramic  semantic segmentation collected from locations all around the world and densely annotated with $19$ classes in accordance to the pinhole camera dataset Cityscapes to enable proper \textsc{Pinhole$\rightarrow$Panoramic} evaluation.
\item We formalize the problem of unsupervised domain adaptation for panoramic segmentation focused on transfer from label-rich pinhole  datasets to DensePASS. 
\item We propose a generic P2PDA framework and investigate different DA modules both in a separate and joint manner, validating their effectiveness with various networks designed for self-driving scene segmentation.
\end{itemize}

\section{RELATED WORK}

\subsection{Semantic Segmentation and Self-attention Modules}

The performance of semantic segmentation models has rapidly improved through the explosive rise of deep learning~\cite{fcn,deeplabv3+,pspnet,denseaspp}.
The first prominent end-to-end  architecture to outperform conventional approaches was the Fully Convolutional Network (FCN)~\cite{fcn} followed by DeepLab~\cite{deeplabv3+}, PSPNet~\cite{pspnet} and DenseASPP~\cite{denseaspp} leveraging atrous convolutions or pyramid pooling.

At the same time, the use of \emph{self-attention modules}~\cite{attention}, which learn to automatically weigh input positions (\textit{i.e.} temporal~\cite{attention} or spatial~\cite{nonlocal}),  increasingly gains interest in the field. 
Such mechanisms are widely used for capturing long-range contextual dependencies which are crucial for dense prediction tasks.
The success of attention mechanisms in  visual recognition~\cite{nonlocal}, leads to their explorations in  some semantic segmentation works focused on both, accuracy-oriented networks~\cite{danet,ranet,ocrnet} and efficiency-oriented networks~\cite{fanet,omnirange,attanet}.
For example, FANet~\cite{fanet} proposed a fast self-attention module for non-local context aggregation aiming efficient segmentation, whereas DANet~\cite{danet} was designed with position and channel attention modules to learn spatial and channel interdependencies. 

We leverage such attention principles to mitigate the domain shift by highlighting regional context and present a cross-domain segmentation framework with \emph{attentional domain adaptation modules}.
We experiment with both, accuracy- and efficient-oriented networks~\cite{danet,fanet,erfnet} as the segmentation architecture, and demonstrate the consistent effectiveness of our adaptation modules for bringing standard semantic segmentation model to panoramic imagery.

\subsection{Semantic Segmentation for Panoramic Images}

Segmentation of panoramic data, which is often captured through distortion-heavy fisheye lenses~\cite{universal,adaptable_deformable} or multiple cameras~\cite{restricted,omnidet}, is especially challenging as it requires eliminating distortions, synchronizing and calibrating the cameras, and fusing the data, which leads to higher latency.
Yang \textit{et al.} introduced the PASS~\cite{pass} and the DS-PASS~\cite{dspass} frameworks which successfully mitigate the effect of distortions by using a panoramic annular lens system but come with a high memory- and computational cost.
This was significantly improved by the OOSS framework~\cite{ooss} through multi-source omni-supervised learning for omnidirectional segmentation.
The latest advancements include frameworks focusing on omni-range contextual dependencies~\cite{omnirange} or leveraging pixel-level contrastive pre-training~\cite{pps}.

All previous frameworks~\cite{pass,omnirange,pps} are developed under the assumption that the labelled training data are implicitly or partially available in our target  domain of panoramic images. 
Since panoramic datasets are comparably small in size, we argue, that \emph{panoramic} segmentation might strongly benefit from the significantly larger datasets available in the domain of \emph{standard} image segmentation.
To achieve this, we look at panoramic segmentation from a \emph{domain adaptation perspective} and introduce the DensePASS dataset covering images with $19$ annotated categories in both, standard- and panoramic domains.
We further introduce a framework for unsupervised domain adaptation for panoramic semantic segmentation, where we combine prominent segmentation approaches with attention-based adaptation modules. 

\subsection{Domain Adaptation for Semantic Segmentation}

Domain adaptation has been recently addressed in the \textit{standard} semantic segmentation, by either (1) optimizing the \textit{source} domain networks on pseudo-labels generated by a model trained in the \textit{source} domain~\cite{pycda,crst}, or (2) leveraging a Generative Adversarial Network (GAN)~\cite{gan} to learn domain translations~\cite{bridging,adaptsegnet,all_about_structure,fcnsinthewild}.
To utilize the significant amount of \textit{source}-\textit{target} similarities in the resulting segmentation masks, Tsai \textit{et al.}~\cite{adaptsegnet} aligned the domains in output space via adversarial learning (AdaptSegNet). 
Chang \textit{et al.}~\cite{all_about_structure} introduced the DISE framework which extracts domain-invariant structure and domain-specific texture information to reduce the  source-target discrepancies.
Attention mechanisms also started gaining attention for domain adaptation with techniques such as attentional transfer~\cite{safe} or multiple cross-domain attention modules for obtaining context dependencies from both local and global perspectives~\cite{contextaware}.

We specifically  focus on domain transfer for \emph{panoramic} semantic segmentation, which differs from the standard \emph{pinhole}  images in several important aspects, such as discontinuous boundaries and distorted objects.
To capture long-range correlations between pixels and semantic regions, we extend AdaptSegNet~\cite{adaptsegnet} with attention-augmented modules in multiple stages and a regional context exchange, leading to a significant adaptation improvement.

\begin{figure*}[!t]
  \centering
  \includegraphics[width= \textwidth]{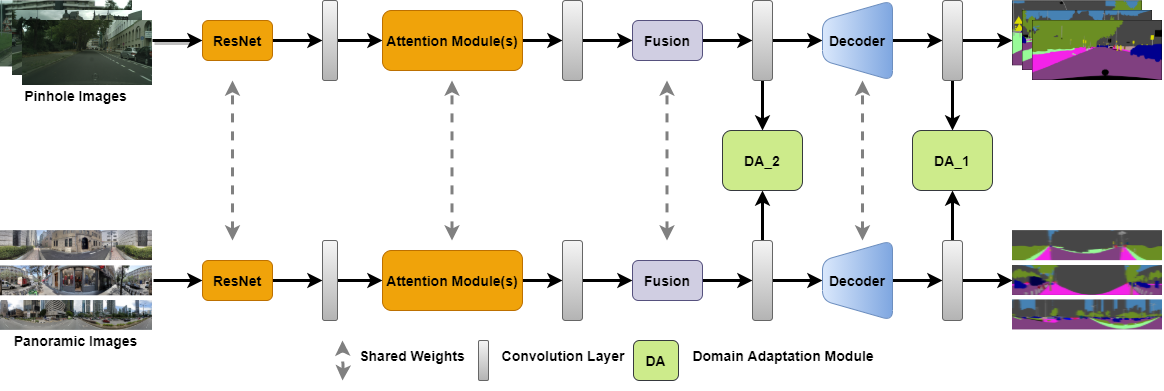}
  \vskip-1ex
  \caption{An overview of the proposed P2PDA framework. 
  Our main backbone is a segmentation network with attention modules. DA\_2 and DA\_1 are different domain adaptation modules we use in multi-levels. 
  Images from pinhole source domain and panoramic target domain will be fed into the segmentation network to generate semantic segmentation predictions. Fusion layers fuse the attended feature maps after the attention modules. 
  Meanwhile, feature maps from the last layer and before the decoder will be inputted into DA\_1 and DA\_2, respectively, to adapt from the pinhole source domain to the panoramic target domain. 
  SDAM can be used in both DA\_1 and DA\_2, ADAM can be used in DA\_2 and RCDAM can be used in DA\_1.}
  \vskip-3ex
  \label{fig.overview}
\end{figure*}

\section{P2PDA: PROPOSED FRAMEWORK}

In this work, we introduce a generic framework for  $360^\circ$  perception of self-driving scenes by learning to adapt semantic segmentation networks from a label-rich source domain of standard pinhole camera images to the unlabelled target domain of panoramic data.
Conceptually, our framework comprises encoder-decoder-based semantic segmentation network and three different building blocks for domain alignment: Segmentation domain adaptation module (SDAM), Attentional domain adaptation module (ADAM) and Regional context domain adaptation module (RCDAM), which we place at two different network stages: after and before the decoder of the segmentation network (denoted as DA\_2 and DA\_1 respectively). 
Next, we  present an overview of the proposed framework (Sec. \ref{sec:overview}) and describe the three integrated domain adaptation modules  in detail (Sec. \ref{sec:da-modules}).

\subsection{Framework Overview}
\label{sec:overview}

Our framework builds on the AdaptSegNet model~\cite{adaptsegnet}, extending it with multiple versions of region- or attention-augmented DA modules placed at different network levels with an overview provided in Fig.~\ref{fig.overview}. 
The main components of our framework are a weight-shared segmentation network \textbf{G} with attention modules and two DA modules (DA\_2 and DA\_1) with corresponding discriminators \textbf{D}.
We denote the source domain images as $I_s$ and target domain images as $I_t$. For simplicity, $I_s$ and $I_t$ also refer to the intermediate feature map representations of the images.

At first, the source domain images $I_s$  are fed into the segmentation network \textbf{G} (also referred to as the generator) to generate prediction results and  the source ground-truth labels are used to compute the segmentation loss $\mathcal{L}_{seg}$.
Next, the corresponding feature maps are passed to the domain adaptation modules in order to close the gap  between the source and target domains at different network levels.
The discriminators are trained with the binary objective to estimate the domain of the input data, so that the discriminator loss $\mathcal{L}_{D}(I_s,I_t)$ is a cross-entropy loss with two classes (panoramic and pinhole). 
The discriminator output of the target domain data $I_t$ is directly used to estimate the adversarial loss $\mathcal{L}_{adv}[\textbf{D}(I_t)]$ for the generator training (alongside with $\mathcal{L}_{seg}$) and is high if the discriminator prediction is correct (so the adversarial loss facilitates generation of segmentation masks in the target domain which successfully ``fool'' the discriminator).
In other words, the discriminators are trained to distinguish between the source and target domains with $\mathcal{L}_{D}(I_s,I_t)$, while the segmentation network $\textbf{G}$ is trained to 1) correctly segment the images from the source domain with $\mathcal{L}_{seg}$, and 2) ``fool'' the discriminator by making the target domain data indistinguishable from the source domain data.
The final loss used to train the generator becomes:

\[ \mathcal{L}_{G}(I_s, I_t) = \lambda_{seg}\mathcal{L}_{seg}(I_s) + \lambda_{adv}\mathcal{L}_{adv}[\textbf{D}(I_t)], \] where $\lambda_{adv}$ and $\lambda_{seg}$ are weights used to balance the domain adaptation and semantic segmentation losses.

\subsection{Domain Adaptation Modules}
\label{sec:da-modules}

We now explain the three integrated DA modules in detail.

\begin{figure}[!t]
  \centering
  \includegraphics[width=0.485\textwidth]{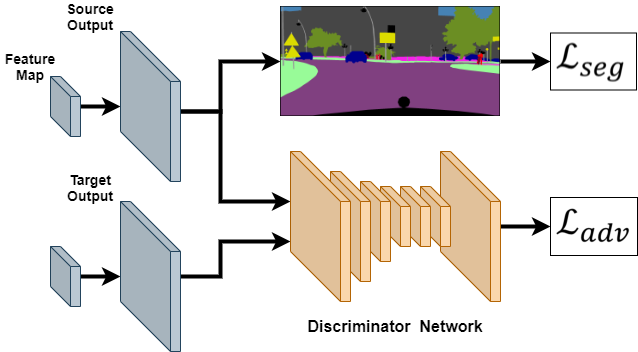}
  \vskip-1ex
  \caption{An overview of the SDAM module.}
  \label{fig.overview.segmentation.da}
  \vskip-3ex
\end{figure}

\noindent  
\textbf{Segmentation domain adaptation module (SDAM).}
Our initial domain adaptation module SDAM is derived from AdaptSegNet and attempts to align the segmentation output of the source and target maps (the module is illustrated in Fig.~\ref{fig.overview.segmentation.da}).
After a segmentation network forward pass with both, an image from the source and target domains ($I_s$ and $I_t$), feature maps of the both representations are used as input to the discriminator \textbf{D} which learns to predict the domain with $\mathcal{L}_{D}(I_s,I_t)$, while the segmentation network $G$ learns to correctly segment the pinhole images with $\mathcal{L}_{seg}(I_s)$ and align the domains with ${L}_{adv}[\textbf{D}(I_t)]$.
The SDAM learns a \textsc{Pinhole$\rightarrow$Panoramic} domain adaptation model at \emph{multiple levels jointly} (\textit{i.e.} integration in DA\_1 and DA\_2).

\begin{figure}[h]
  \centering
  \includegraphics[width=0.485\textwidth]{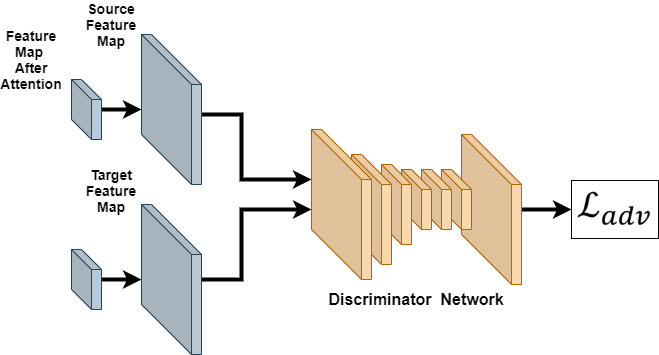}
  \vskip-1ex
  \caption{An overview of the ADAM module.}
  \vskip-2ex
  \label{fig.overview.attention.da}
\end{figure}

\noindent  
\textbf{Attentional domain adaptation module (ADAM).}
To detect and effectively utilize the significant amount of pinhole-panoramic correspondences at both, local and global scale, we design \emph{ADAM}, an \emph{attentional} domain adaptation module (overview in Fig.~\ref{fig.overview.attention.da}).
ADAM differs from SDAM as it leverages the attention mechanism to learn an optimal weighting scheme for the features used as the discriminator input. 
By doing this, ADAM enables direct information exchange among all pixels, mitigating the influence of discrepancy in positional priors and local distortions.
Relevant portions of the feature maps of both, $I_s$ and $I_t$ inputs are magnified through the  attention and the \emph{re-weighted} source and target representations are both used to optimize the corresponding discriminator \textbf{D} with adversarial loss.
Since ADAM  operates on attended feature maps with long-range contextual dependency aggregation, it is used in DA\_2 only.

\begin{figure}[!t]
  \centering
  \includegraphics[width=0.485\textwidth]{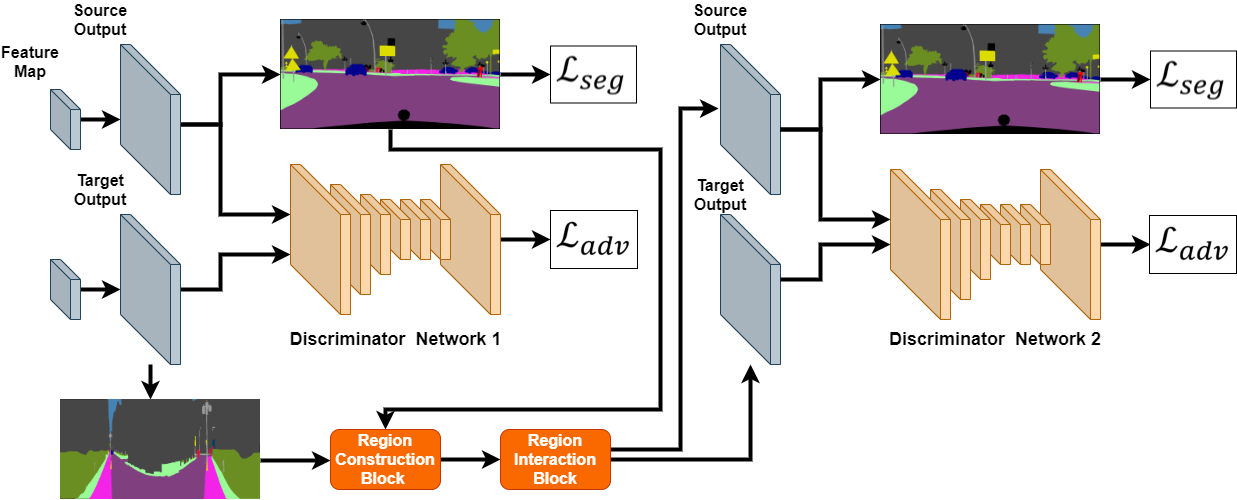}
  \vskip-1ex
  \caption{An overview of the RCDAM module.}
  \label{fig.overview.regional.da}
  \vskip-3ex
\end{figure}

\noindent  
\textbf{Regional context domain adaptation module (RCDAM).}
Next, we focus on \emph{region relationship of the panoramic images}. 
Inspired by RANet~\cite{ranet}, we design the RCDAM module to configure the information flow  between different regions and within the same region, as illustrated in Fig.~\ref{fig.overview.regional.da}.

RCDAM follows a hierarchical adversarial learning scheme with two-stage discriminators, where the first stage is identical to the previously described SDAM.
The second stage covers two blocks: a Region Construction Block (RCB) and a Region Interaction Block (RIB) first introduced in RANet.
The inputs to this stage are the feature maps of $I_s$ and $I_t$ after a segmentation network forward pass. 

\begin{figure}[h]
  \centering
  \includegraphics[width=0.485\textwidth]{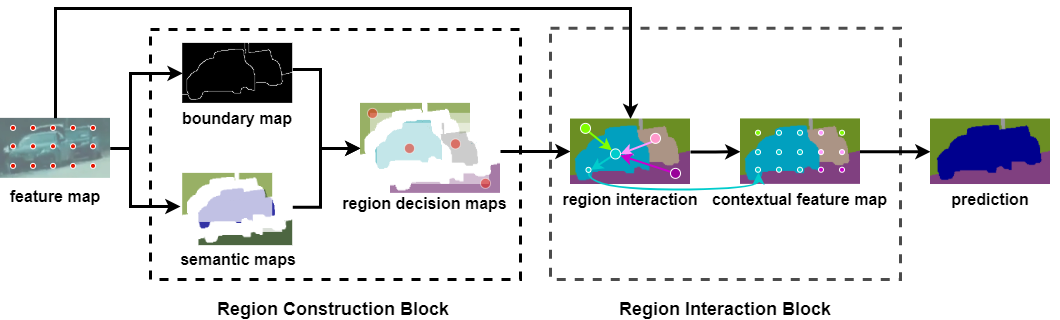}
  \vskip-1ex
  \caption{Diagram of Region Construction and Region Interaction Blocks.}
  \label{fig.overview.blocks.da}
  \vskip-2ex
\end{figure}
Fig.~\ref{fig.overview.blocks.da} gives a detailed overview of the RCB and RIB building blocks.
RCB computes a  boundary map and semantic maps as in~\cite{ranet} to link each pixel (the red dot in Fig.~\ref{fig.overview.blocks.da}) to different semantic regions. 
Then, RCB conducts a region decision map with the representative pixels (the larger dots). 
After that, RIB selects representative pixels (the larger dots in color) for each region and summarizes information from the pixels (the smaller dots) in the same region into representative pixels. 
Next, RIB aggregates the information from representative pixels in other regions to form the global contextual representations which are afterwards propagated to all pixels in the image. 
At last, regional context is sufficiently exchanged and new regional context-driven feature maps are built to generate final prediction outputs, which will be used to perform the same operations as in the first stage. 
It is worth noting that, the second stage is used to help improve the segmentation result in the first stage, which can make the result more compact and helps the adaptation between the pinhole domain and the panoramic domain by exchanging regional context. 
Thereby, the final output is from the first stage, which means this domain adaptation module does not change the original architecture of the segmentation network and can be used for networks without attention layers.

\section{NEW BENCHMARK AND EXPERIMENTS}

To  evaluate the  idea  of  panoramic  segmentation through domain adaptation from  standard pinhole camera images, we conduct extensive experiments with different variants of our P2PDA framework.
First, we introduce the novel DensePASS dataset for dense \emph{panoramic} segmentation of driving scenes annotated in accordance to the \emph{pinhole} camera  dataset Cityscapes (Sec. \ref{sec:dataset}). 
Then, we  quantitatively validate how well the P2PDA framework can handle the \textsc{Pinhole$\rightarrow$Panoramic} transfer and conduct extensive ablation studies for different versions of DA modules and two  segmentation networks: the speed-oriented FANet~\cite{fanet} (Sec. \ref{sec:fanet}) and accuracy-oriented  DANet~\cite{danet} (Sec. \ref{sec:danet}).
We further benchmark our framework against more than $15$ state-of-the-art segmentation models (Sec. \ref{sec:soa}) and examine the impact of expanding the training data  with another source dataset (Sec. \ref{sec:wilddash}).
Finally, we showcase multiple qualitative examples of the achieved results (Sec. \ref{sec:qualitative}).
We adopt Mean Intersection over Union (IoU) as our main evaluation metric.

\begin{table*}[!t]
\renewcommand\arraystretch{1.2}
      \footnotesize
      \setlength{\tabcolsep}{4pt}
      \begin{center}
      \caption{Per-class results on DensePASS. We use FANet~\cite{fanet} as the segmentation network and set different domain adaptation modules for DA\_2 and DA\_1 to test our methods on DensePASS with the size of input $2048{\times}400$. S represents the segmentation domain adaptation module, A represents the attentional domain adaptation module and S+A represents a combination of S and A. The first line is the Cityscapes source-only result without adaptation.}
      \label{tab:fanet}
      \vskip-1ex
      \begin{adjustbox}{max width=\textwidth}
          \begin{tabular}{ l | c |c | c c c c c c c c c c c c c c c c c c c c}
              \toprule[1pt]
         Methods & DA\_2 & DA\_1 &  \rotatebox{90}{Mean IoU} &  \rotatebox{90}{road} &  \rotatebox{90}{sidewalk} &  \rotatebox{90}{building} & \rotatebox{90}{ wall} &  \rotatebox{90}{fence} &  \rotatebox{90}{pole} & \rotatebox{90}{traffic light} &  \rotatebox{90}{traffic sign}&  \rotatebox{90}{vegetation} &  \rotatebox{90}{terrain} &  \rotatebox{90}{sky} & \rotatebox{90}{person} &  \rotatebox{90}{rider} & \rotatebox{90}{car} &  \rotatebox{90}{truck}& \rotatebox{90}{ bus}& \rotatebox{90}{ train}& \rotatebox{90}{ motorcycle}&  \rotatebox{90}{bicycle}\\
        \hline
        \hline
        FANet & - & - & 26.90 & \textbf{62.98} & 10.64 & 72.41 & 7.80 & 20.74 & 11.77 & 6.85 & 3.75 & 68.11 & \textbf{21.56} & \textbf{87.00} & 23.73 & 5.33 & 49.61 & 10.65 & \textbf{0.54} & 16.76 & 24.15 & 6.62 \\
        FANet & S & S & 32.17 & 62.16 & 16.85 & 78.78 & 13.67 & 24.07 & 19.72 & 11.42 & 9.68 & 71.42 & 18.22 & 85.72 & 32.66 & \textbf{11.75} & 54.34 & \textbf{17.61} & 0.00 & 41.52 & \textbf{29.30} & 12.30 \\
        FANet & A & S & 32.67 & 62.28 & 16.86 & 79.99 & \textbf{17.64} & 23.96 & \textbf{19.78} & \textbf{12.33} & 9.58 & 72.01 & 19.29 & 85.91 & 32.85 & 11.03 & \textbf{55.75} & 15.38 & 0.38 & 43.53 & 29.19 & \textbf{12.95} \\
        FANet & S+A & S & \textbf{33.05} & 61.74 & \textbf{17.70} & \textbf{80.07} & 16.38 & \textbf{24.64} & 19.61 & 12.04 & \textbf{9.79} & \textbf{72.27} & 17.94 & 86.31 & \textbf{33.17} & 11.47 & 55.18 & 15.61 & 0.04 & \textbf{52.55} & 28.68 & 12.82 \\
          \bottomrule[1pt]
          \end{tabular}
      \end{adjustbox}
      \end{center}
  \vskip-4ex
  \end{table*}

\subsection{Dataset and Experimental Settings}
\label{sec:dataset}

\noindent
\textbf{Source dataset (pinhole).} 
We use Cityscapes~\cite{cityscapes} as our label-rich source dataset providing a large amount of annotated pinhole camera images.
Cityscapes covers $2979$ training- and $500$ validation examples captured in $50$ different European cities and annotated with $19$ categories. 
We use the $2979$ training samples as our source training data. 

\noindent    
\textbf{DensePASS target dataset (panoramic).} 
Since no established segmentation benchmarks target the \textsc{Pinhole$\rightarrow$Panoramic} transfer and previous panoramic test-beds cover a very small number of classes~\cite{pass}\cite{omnirange}, we collect DensePASS -- a novel densely annotated dataset for panoramic segmentation of driving scenes.
While DensePASS could also be used for the the same-domain panormaic segmentation, it is specifically built with the \textsc{Pinhole$\rightarrow$Panoramic} transfer in mind, so that the test data is annotated with $19$ categories present in the pinhole camera dataset Cityscapes. 
To facilitate the unsupervised domain adaptation task, DensePASS covers both, labelled data ($100$ panoramic images used for testing) and unlabelled training data ($2000$ panoramic images used for the domain transfer optimization).
Panoramic images ($2048{\times}400$ resolution) are collected using Google Street View, spanning images from different continents ($25$ different cities for testing and $40$ for training).

\noindent
\textbf{Training settings.}
We use stochastic gradient descent (initial learning rate of $1e-5$, momentum of $0.9$ and a decay of $5e-4$) for optimization of the segmentation network \textbf{G} and the Adam optimizer for  discriminators \textbf{D} (initial learning rate set to $4e-6$).
For both optimizers, the learning rate is decreased polynomially where the learning rate is multiplied by $(1-\frac{iter}{max\_iter})^{0.9}$ after each iteration, where max\_iter is set to $200000$ with a batch-size of $2$. 
The loss balancing weights, $\lambda_{adv}$ and $\lambda_{seg}$ are set to $0.001$ and $1.0$ for DA\_1, and $0.0002$ and $0.1$ for DA\_2.
In the RCDAM module, $\lambda_{seg}$ is set to $1.5$ for the prediction result before RCB. 
During training, pinhole images are resized to $1280{\times}720$ while panoramic data is kept at the $2048{\times}400$ resolution.

\subsection{Ablation Studies for Segmentation Network with FANet}
\label{sec:fanet}
We first consider FANet~\cite{fanet}, a lightweight speed-oriented network, as the segmentation model and investigate different combinations of domain adaptation modules for DA\_2 and DA\_1. 
As shown in Table~\ref{tab:fanet}, before adaptation, FANet yields a mean IoU of $26.90\%$ indicating large room for improvement in cross-domain generalization.
Our framework improves the result to $32.17\%$ by using the SDAM module in both DA\_2 and DA\_1 ($5.27\%$ gain).
Integrating the attentional ADAM module also leads to a considerable boost ($32.67\%$ IOU, a $5.77\%$ gain over the  the source-only baseline). 
A combination of the SDAM and ADAM modules in DA\_2 yields the best recognition result of $33.05\%$ IoU.

\subsection{Ablation Studies for Segmentation Network with DANet}
\label{sec:danet}

\begin{table*}[!t]
\renewcommand\arraystretch{1.2}
      \footnotesize
      \setlength{\tabcolsep}{4pt}
      \begin{center}
      \caption{Per-class results on DensePASS. We use DANet~\cite{danet} as the segmentation network and set different domain adaptation modules for DA\_2 and DA\_1 to test our methods on DensePASS with the size of input $2048{\times}400$. S represents the segmentation domain adaptation module, A represents the attentional domain adaptation module, R represents the regional context domain adaptation module and S+A represents a combination of S and A. The first line is the Cityscapes source-only result without adaptation. * denotes further adding source images from WildDash to complement Cityscapes.}
      \label{tab:danet}
      \vskip-1ex
      \begin{adjustbox}{max width=\textwidth}
          \begin{tabular}{ l | c |c | c c c c c c c c c c c c c c c c c c c c}
              \toprule[1pt]
         Methods & DA\_2 & DA\_1 &  \rotatebox{90}{Mean IoU} &  \rotatebox{90}{road} &  \rotatebox{90}{sidewalk} &  \rotatebox{90}{building} & \rotatebox{90}{ wall} &  \rotatebox{90}{fence} &  \rotatebox{90}{pole} & \rotatebox{90}{traffic light} &  \rotatebox{90}{traffic sign}&  \rotatebox{90}{vegetation} &  \rotatebox{90}{terrain} &  \rotatebox{90}{sky} & \rotatebox{90}{person} &  \rotatebox{90}{rider} & \rotatebox{90}{car} &  \rotatebox{90}{truck}& \rotatebox{90}{ bus}& \rotatebox{90}{ train}& \rotatebox{90}{ motorcycle}&  \rotatebox{90}{bicycle}\\
        \hline
        \hline
            DANet & - & - & 28.50 & \textbf{70.68} & 8.30 & 75.80 & 9.49 & 21.64 & \textbf{15.91} & 5.85 & 9.26 & \textbf{71.08} & \textbf{31.50} & \textbf{85.13} & 6.55 & 1.68 & 55.48 & 24.91 & 30.22 & 0.52 & 0.53 & 17.00 \\
            DANet & S & S & 38.51 & 61.78 & 21.11 & 74.59 & 22.59 & 29.93 & 14.79 & 15.00 & 10.17 & 66.94 & 19.03 & 82.57 & 31.03 & 21.24 & 53.26 & \textbf{54.67} & 37.77 & 39.40 & 43.84 & 31.95 \\
            DANet & A & S & 39.16 & 61.34 & 20.71 & 76.52 & 20.53 & 30.03 & 14.19 & 15.69 & 10.09 & 68.60 & 18.84 & 82.08 & 33.16 & \textbf{21.75} & \textbf{57.68} & 53.88 & 40.33 & 41.47 & 46.11 & 31.00 \\
            DANet & S+A & S & 39.28 & 62.43 & 21.89 & 76.22 & 21.42 & 30.54 & 14.85 & 14.10 & 9.76 & 69.07 & 19.94 & 82.84 & \textbf{34.56} & 19.30 & 56.51 & 53.04 & 42.51 & 39.47 & 45.71 & 32.09 \\
            DANet & S & R & 39.46 & 62.75 & 23.17 & \textbf{76.65} & 23.90 & \textbf{30.82} & 14.84 & \textbf{18.44} & 10.09 & 69.10 & 17.60 & 82.78 & 33.51 & 21.53 & 55.97 & 51.78 & 41.77 & 36.90 & 46.11 & \textbf{32.12} \\
            DANet & S+A & R & \textbf{39.76} & 63.11 & \textbf{24.63} & 76.17 & \textbf{25.03} & 30.56 & 13.68 & 15.68 & \textbf{10.53} & 67.31 & 22.41 & 80.15 & 32.95 & 21.11 & 54.39 & 53.51 & \textbf{43.64} & \textbf{42.20} & \textbf{46.71} & 31.66 \\
            \midrule
            DANet* & S & R & 41.35 & \textbf{68.38} & \textbf{37.26} & \textbf{75.51} & 26.28 & 31.81 & 15.62 & 8.99 & \textbf{10.33} & \textbf{66.22} & \textbf{31.74} & \textbf{80.68} & 33.69 & \textbf{16.81} & \textbf{64.81} & 47.67 & 28.05 & 61.81 & 44.92 & \textbf{34.98} \\
            DANet* & S+A & R & \textbf{42.47} & 67.47 & 30.16 & 75.27 & \textbf{30.26} & \textbf{37.50} & \textbf{16.19} & \textbf{9.35} & 9.78 & 63.14 & 30.44 & 77.07 & \textbf{34.82} & 15.24 & 64.33 & \textbf{53.70} & \textbf{43.33} & \textbf{71.57} & \textbf{46.80} & 30.47 \\
          \bottomrule[1pt]
          \end{tabular}
      \end{adjustbox}
      \end{center}
      \vskip-3ex
  \end{table*}
 
Next, we experiment with a heavier and more accuracy-oriented segmentation network DANet~\cite{danet} (results provided in Table~\ref{tab:danet}).
The native source-trained DANet achieves a mean IoU of only $28.50\%$, highlighting the sensitivity of modern segmentation networks to the \textsc{Pinhole$\rightarrow$Panoramic} domain shift.
The performance is strongly improved ($10.01\%$ boost) through SDAM modules placed in DA\_2 and DA\_1, achieving $38.51\%$ in mean IoU.
Similarly, using the ADAM module in DA\_2 yields a result of $39.16\%$ (a $10.66\%$ improvement over the source-only baseline). 
Combining both the SDAM and ADAM modules again slightly improves the performance ($39.28\%$ mean IoU).

We further explore the use of the RCDAM module in DA\_1, yielding  $39.46\%$ mIoU ($10.96\%$ boost over the baseline).
The best performance of $39.76\%$ ($11.26\%$ boost with respect to the original segmentation network) is achieved by combining all three modules: SDAM and ADAM in DA\_2 and RCDAM in DA\_1.
Overall, our experiments showcase that  direct cross-domain semantic segmentation is a hard task and P2PDA framework with attentional and regional domain adaptation modules clearly helps to close the domain gap.

\begin{table}[!t]
\caption{Performance of models on Cityscapes and DensePASS.}
\vskip -1ex
\label{tab:domain_gap}
\centering
\resizebox{\columnwidth}{!}{
\begin{tabular}{@{}llrrc@{}}
\toprule
\textbf{Network} & \textbf{Backbone} & \textbf{Cityscapes} & \textbf{DensePASS} & \textbf{mIoU Gap} \\ \midrule \midrule
DeepLabV3+~\cite{deeplabv3+} & ResNet-18    & 76.8  &  25.6   & -51.2 \\ 
OCRNet~\cite{ocrnet}            & HRNetV2p-W18s & 77.1 & 25.9   & -51.2 \\
Fast-SCNN~\cite{fastscnn}     & Fast-SCNN	   & 69.1  &  24.6   & -44.5 \\
\midrule
DeepLabV3+~\cite{deeplabv3+} & ResNet-50    & 80.1  &  29.0  & -51.1 \\ 
PSPNet~\cite{pspnet}            & ResNet-50    & 78.6  &  29.5  & -49.1 \\
DNL~\cite{dnl}                   & ResNet-50    & 79.3  & 28.7  & -50.6 \\
Semantic FPN~\cite{panopticfpn} & ResNet-50 & 74.5 &  29.9   & -44.6 \\
OCRNet~\cite{ocrnet}            & HRNetV2p-W18 & 78.6  & 30.8  & -47.8 \\ 
\midrule
DeepLabV3+~\cite{deeplabv3+} & ResNet-101   & 80.9  &  32.5  & -48.4 \\
PSPNet~\cite{pspnet}            & ResNet-101   & 79.8  &  30.4  & -49.4 \\
DNL~\cite{dnl}                   & ResNet-101   & 80.4  & 32.1  & -48.3 \\
Semantic FPN~\cite{panopticfpn} & ResNet-101 & 75.8 &  28.8  & -47.0 \\
ResNeSt~\cite{resnest}         & ResNeSt-101  & 79.6  &  28.8   & -50.8 \\
OCRNet~\cite{ocrnet}            & HRNetV2p-W48 & 80.7  &  32.8  & -47.9 \\
\midrule \midrule
ERFNet~\cite{erfnet} & ERFNet    & 72.1  & 16.7 &  -55.4  \\ 
ERFNet-P2PDA (Ours) & ERFNet & 72.1 & 34.1 & -38.0 \\
FANet~\cite{fanet} & ResNet-34 & 71.3 & 26.9 & -44.4 \\
FANet-P2PDA (Ours) & ResNet-34 & 71.3 & 33.1 & -38.2 \\
DANet~\cite{danet} & ResNet-50 & 79.3  &  28.5  & -50.8 \\
DANet-P2PDA  (Ours) & ResNet-50 & 79.3 & 39.8 & -39.5 \\
\bottomrule
\end{tabular}}
\vskip-3ex
\end{table}

\subsection{Benchmarking and Comparison with the State-of-the-Art}
\label{sec:soa}

Our previous experiments compared different P2PDA framework configurations with each other and  the native segmentation network.
Next, we aim to quantify the domain gap between the pinhole- and  panoramic images and extend our evaluation with over $15$ off-the-shelf segmentation models trained on Cityscapes and evaluated on both, Cityscapes (no domain shift) and the panoramic DensePASS images.\footnote[1]{For a fair comparison, model weights are provided by the same framework MMSegmentation: https://github.com/open-mmlab/mmsegmentation}
Table~\ref{tab:domain_gap} summarizes our results.
It is evident, that modern segmentation models trained on pinhole camera images struggle with generalization to  panormic data,  with performance degrading by $\sim50\%$ as we move from the standard \textsc{Pinhole$\rightarrow$Pinhole} setting to the \textsc{Pinhole$\rightarrow$Panoramic} evaluation. 
We also  verify our P2PDA domain adaptation strategy with three different  segmentation models (bottom part of Table~\ref{tab:domain_gap}).
Our P2PDA strategy with regional and attentional context exchange improves the \textsc{Pinhole$\rightarrow$Panoramic} results by a large margin (mIoU gains of $17.4\%$, $6.2\%$ and $11.3\%$ for ERFNet, FANet and DANet, respectively).
Our experiments provide encouraging evidence that P2PDA can be successfully deployed for cross-domain $360^\circ$ understanding of driving scenes.

\subsection{Complementing the Cityscapes Source with WildDash}
\label{sec:wilddash}

Our next area of investigation is the impact of expanding the source  data with a more complex dataset, since DensePASS contains highly composite scenes due to larger FoV, while Cityscapes is  large but relatively simple.
To achieve this, we leverage the WildDash dataset~\cite{wilddash} with  $4256$ pinhole images, pixel-level annotations and more unstructured surroundings.
As shown in the last two rows of Table~\ref{tab:danet}, we obtain better mIoU with the expanded training set, achieving $41.35\%$ and $42.47\%$ with different P2PDA variants.
Interestingly, the IoUs of \emph{road}, \emph{sidewalk}, \emph{terrain}, \emph{car}, and \emph{train} are significantly improved, which we link to the strong positional priors of these categories in structured urban scenes, while DensePASS and WildDash environment is more chaotic and unconstrained.
Furthermore, direct comparison of the two  P2PDA versions demonstrates the effectiveness of the attention-augmented adaptation strategy.

\subsection{Qualitative Analysis}
\label{sec:qualitative}

\begin{figure*}[!t]
  \centering
  \includegraphics[width= \textwidth]{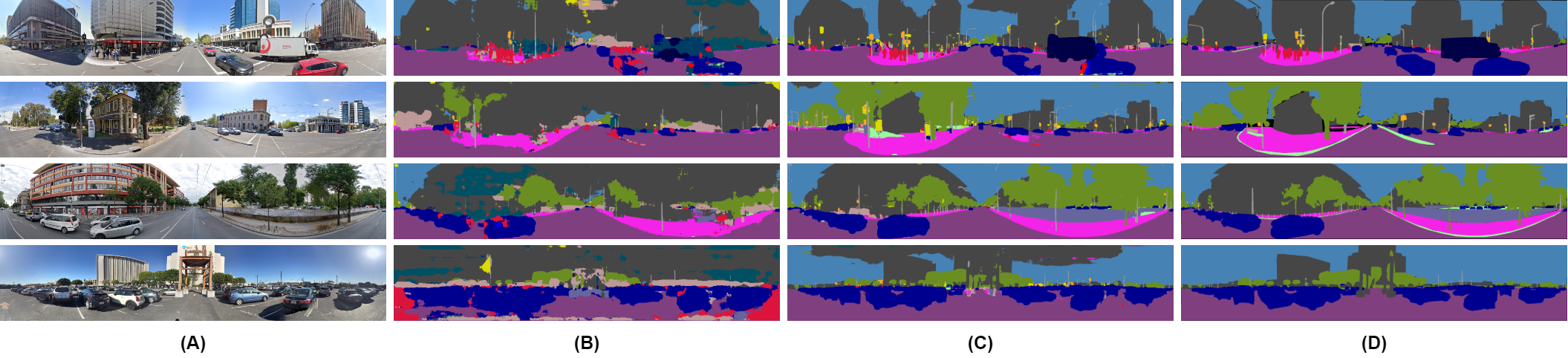}
  \vskip-2ex
  \caption{Qualitative examples of semantic segmentation on $360^\circ$ panoramic images. 
  For each row, (A) is the original input panoramic image, 
  (B) is the segmentation result without any adaptation, 
  (C) is the segmentation result with our proposed P2PDA adaptation framework, 
  and (D) is the ground truth.}
  \vskip-3ex
  \label{fig.comparision}
\end{figure*}

In our final study, we showcase multiple examples of representative qualitative results in  Fig.~\ref{fig.comparision}.
The segmentation boundaries of regions such as \emph{sky}, \emph{building}, and \emph{vegetation} are clearly improved through the P2PDA strategy in every case, while \emph{sidewalk} segmentation clearly benefits from domain adaptation in the second row example.
At the same time, some misclassified categories are corrected after adaptation (\textit{e.g.}, \emph{car} and \emph{truck} in all examples). 
There is also more clarity as it comes to detailed segmentation of small objects, such as \emph{traffic light} and \emph{traffic sign} in the first and the second row, as well as \emph{pole} in all rows. 
These qualitative examples consistently confirm the conclusions of our quantitative evaluation, highlighting the benefits of the proposed P2PDA strategy for $360^\circ$ self-driving scene understanding through domain adaptation from pinhole camera data. 

\section{CONCLUSION}

Semantic scene understanding is vital for autonomous driving but requires models which can deal with changes in data distribution.
In  this  work,  we introduced the new task of cross-domain semantic segmentation for panoramic driving scenes,  which extends the standard panoramic segmentation with the premise of the training data originating from a different domain (\textit{e.g.} pinhole camera images).
First, we formulate the problem of unsupervised domain adaptation for panoramic  segmentation  and introduce the novel DensePASS dataset which we use to study the \textsc{Pinhole$\rightarrow$Panoramic} transfer.
To meet the challenge of domain divergence, we developed a generic framework enhancing conventional segmentation algorithms with different domain adaptation modules.
While our experiments demonstrate that cross-domain panoramic segmentation task is difficult for modern algorithms, our proposed domain-agnostic framework with attention-based domain adaptation modules consistently improves the results.
Our dataset will be  publicly  released  upon  publication  and  we  believe  that DensePASS has strong  potential to motivate the needed development of \emph{generalizable} semantic segmentation  models.

\bibliographystyle{ieeetr}
\bibliography{bib}

\end{document}